\pdfoutput=1

\documentclass[11pt]{article}

\usepackage{ACL2023}

\usepackage{times}
\usepackage{latexsym}

\usepackage{graphicx}
\usepackage{multicol}
\usepackage{multirow}
\usepackage{booktabs}
\usepackage{subfigure}
\usepackage{amsmath}
\usepackage{amssymb}
\usepackage{hyperref}

\usepackage[T1]{fontenc}

\usepackage[utf8]{inputenc}

\usepackage{microtype}

\usepackage{inconsolata}

\title{Dual-Alignment Pre-training for Cross-lingual Sentence Embedding}

\author{Ziheng Li\textsuperscript{1,}\thanks{\ \ Work done during internship at Microsoft.}~,
Shaohan Huang\textsuperscript{2},
Zihan Zhang\textsuperscript{2},
Zhi-Hong Deng\textsuperscript{1,}\thanks{\ \ Corresponding Author.}~,
Qiang Lou\textsuperscript{2}, \\
{\bf Haizhen Huang\textsuperscript{2},
Jian Jiao\textsuperscript{2},
Furu Wei\textsuperscript{2},
Weiwei Deng\textsuperscript{2},
Qi Zhang\textsuperscript{2}} \\
\textsuperscript{1}School of Intelligence Science and Technology, Peking University, Beijing, China \\
\textsuperscript{2}Microsoft Corporation \\
  \texttt{\small\{liziheng,zhdeng\}@pku.edu.cn}\\
  \texttt{\small\{shaohanh, zihzha, qilou, hhuang, jiajia, fuwei, dedeng, qizhang\}@microsoft.com}\\}

\begin{document}
\maketitle
\begin{abstract}
Recent studies have shown that dual encoder models trained with the sentence-level translation ranking task are effective methods for cross-lingual sentence embedding. However, our research indicates that token-level alignment is also crucial in multilingual scenarios, which has not been fully explored previously. Based on our findings, we propose a dual-alignment pre-training (DAP) framework for cross-lingual sentence embedding that incorporates both sentence-level and token-level alignment. To achieve this, we introduce a novel representation translation learning (RTL) task, where the model learns to use one-side contextualized token representation to reconstruct its translation counterpart. This reconstruction objective encourages the model to embed translation information into the token representation. Compared to other token-level alignment methods such as translation language modeling, RTL is more suitable for dual encoder architectures and is computationally efficient. Extensive experiments on three sentence-level cross-lingual benchmarks demonstrate that our approach can significantly improve sentence embedding. Our code is available at \url{https://github.com/ChillingDream/DAP}.
\end{abstract}

\section{Introduction}
Cross-lingual sentence embedding encodes multilingual texts into a single unified vector space for a variety of Natural Language Processing (NLP) tasks, including cross-lingual sentence retrieval~\citep{laser} and cross-lingual natural language inference~\citep{conneau_xnli_2018}. The text sequences can be efficiently retrieved and compared using the inner product between their dense representation.

\begin{figure}[t]
    \centering
    \subfigure[Sentence Alignment.]{
        \label{fig:wo_align}
        \includegraphics[width=0.46\linewidth]{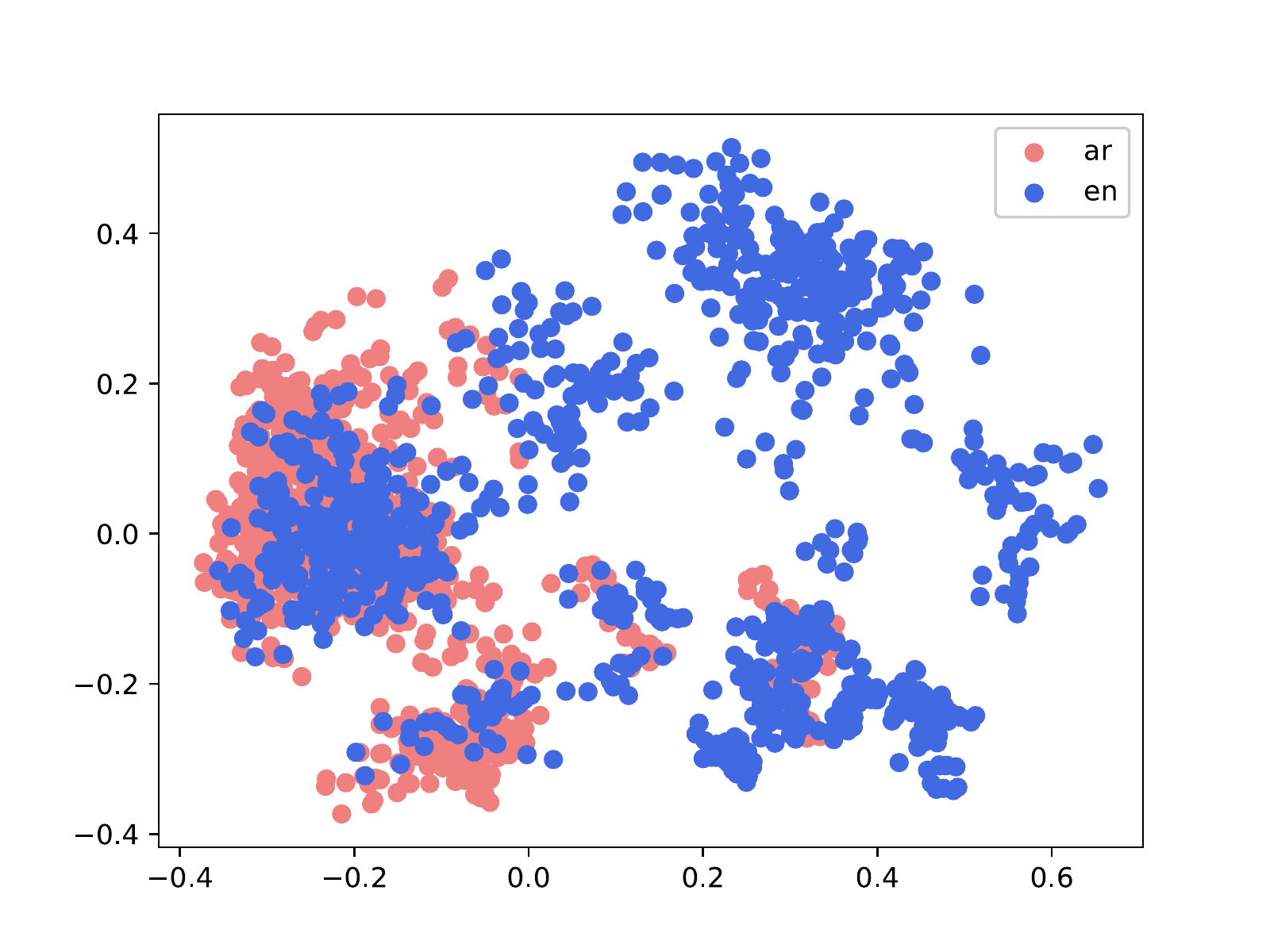}
    }
    \subfigure[Dual Alignment.]{
        \label{fig:rtl_align}
        \includegraphics[width=0.46\linewidth]{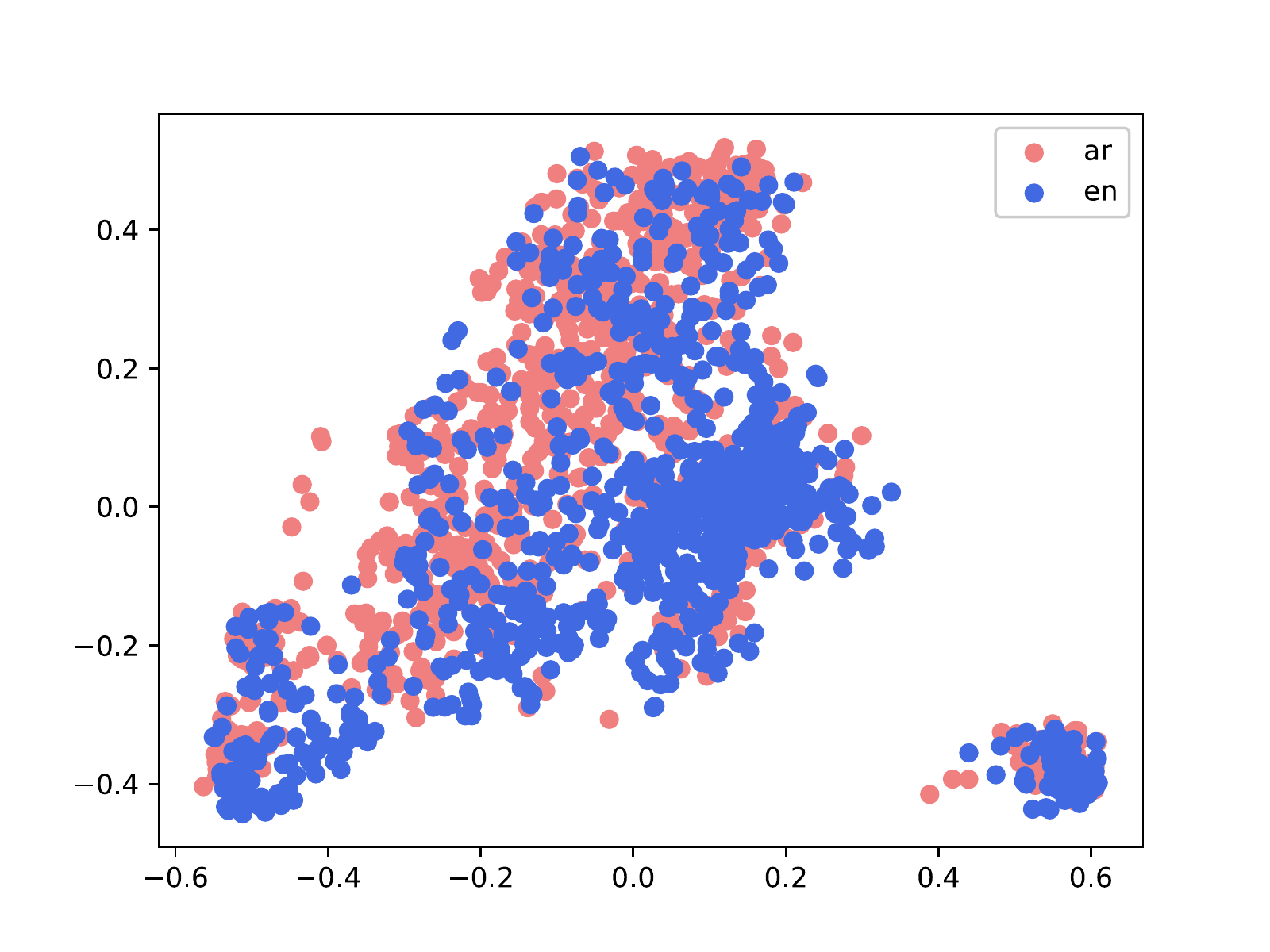}
    }
    \caption{Visualization of token representations of 100 Tatoeba sentence pairs from Arabic and English. The high-dimensional vectors are projected onto a 2D space by Principle Component Analysis. We show the results of two models fine-tuned from multilingual BERT. The model shown in Figure~\ref{fig:wo_align} only fine-tunes with the translation ranking task, resulting in large misaligned areas. This misalignment can be effectively eliminated by the proposed RTL methods as shown in~\ref{fig:rtl_align}.}
    \label{fig:algin_vis}
\end{figure}

The task of sentence embedding now heavily depends on pre-trained language models~\cite{devlin_bert_2019, conneau_cross-lingual_2019, conneau_emerging_2020, conneau_unsupervised_2020}. By fine-tuning the CLS token of the pre-trained model, they encode the input text sequence into a single vector representation. Recent research has shown that using the translation ranking task in combination with a dual pre-trained encoder can result in superior sentence embeddings~\citep{yang_improving_2019,chidambaram_learning_2019,yang_universal_2021,chi-etal-2021-infoxlm,labse}. The purpose of fine-tuning the CLS token is to learn sentence-level alignment and to compress the entire sentence's information into the CLS token. This method makes the CLS tokens of semantically relevant sentences have larger inner products. However, token-level alignment in multilingual scenarios is also crucial, and the fine-grained alignment task in cross-lingual sentence embedding has not been fully explored. As shown in Figure~\ref{fig:algin_vis}, we visualize the token representation similarities between a pair of parallel corpora. Training for an objective solely with regard to CLS token causes the token representations to disperse across the embedding space.

Based on our observations, we propose an efficient dual-alignment pre-training (DAP) framework for cross-lingual sentence embedding. The embedding model is trained towards both sentence-level alignment and token-level alignment. Previous cross-lingual pre-training studies~\citep{chi-etal-2021-infoxlm,labse} employ translation language modeling (TLM) to achieve token alignment. In this paper,  we introduce a novel representation translation learning (RTL) method that reconstructs the entire English input based on the token representations of parallel non-English sentences using a transformer model. By optimizing the RTL objective, the model learns to embed the information of English sentences into the representation of its non-English counterpart. Unlike TLM, computing RTL only needs one-side self-contextualized representation and does not involve extra feedforward propagation. We train our model on public corpora and evaluate it on three cross-lingual tasks: bitext retrieval, bitext mining, and cross-lingual natural language inference. Our results demonstrate DAP can effectively improve cross-lingual sentence embedding.
\begin{figure*}[t]
    \centering
    \includegraphics[width=\linewidth]{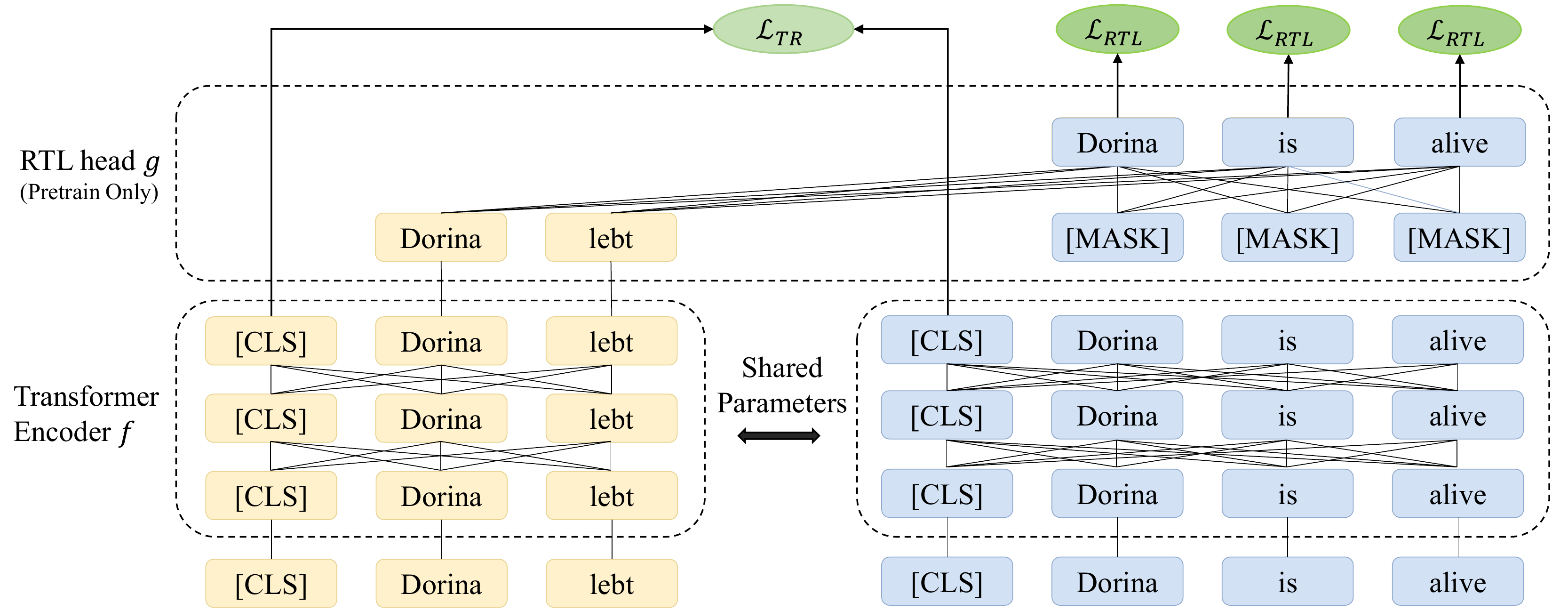}
    \caption{Workflow of the dual-alignment pre-training framework. We encode the bitext pair in dual encoder manner with a shared 12-layer transformer encoder and compute translation ranking loss and representation translation loss using sentence representation and token representations respectively.}
    \label{fig:arch}
\end{figure*}

Our contributions are summarized as follows:
\begin{itemize}
    \item We propose a novel cross-lingual pre-training framework DAP for sentence-level tasks, achieving both sentence-level and token-level alignment by representation translation learning, which is more suitable for dual encoders and computationally efficient compared with previous alignment methods.
    \item Extensive experiments on three cross-lingual tasks demonstrate DAP significantly improves sentence embedding.
    \item We train a model on a moderate-size dataset and find its performance comparable with that of the large-scale state-of-the-art pre-trained model.
\end{itemize}

\section{Related Work}
\subsection{Cross-lingual Pre-training}
Following the success of BERT for English~\citep{devlin_bert_2019}, multilingual BERT comes out by building a shared multilingual vocabulary and training on multiple monolingual corpora with the masked language modeling (MLM) objective. XLM~\citep{conneau_cross-lingual_2019} proposes a translation language modeling (TLM) task which is the extension of MLM to bitext corpora, so that the model can learn the cross-lingual alignment from translation pairs. Unicoder~\citep{huang_unicoder_2019} introduces three bitext pre-training tasks to help the model capture cross-lingual information from more perspectives. XLM-R~\citep{conneau_unsupervised_2020} scales up the amount of monolingual data and training time. They achieve better performance than previous works without using parallel corpora.
\subsection{Sentence Embedding}
The dual encoder architecture is first proposed by~\citet{guo_effective_2018}. They encode the source and target sentences to a unified embedding space, respectively, and compute the similarity score using inner product. The model is trained under a translation ranking task to make the model score higher for translation pairs than the negative examples. \citet{yang_improving_2019} enhances the dual encoder by additive margin softmax, which further enlarges the distance between negative pairs. Based on additive margin softmax, LaBSE~\citep{labse} combines the translation ranking task with MLM task and TLM task and trains on a larger corpus. InfoXLM~\cite{chi-etal-2021-infoxlm} interprets the MLM, TLM and translation ranking task used in cross-lingual pre-training in a unified information-theoretic framework, based on which they propose cross-lingual contrastive learning to maximize sentence-level mutual information.

\section{Method}
\subsection{Preliminaries}
\paragraph{Transformer Encoder}
Transformer encoder has been widely adopted in modern language models~\citep{vaswani_attention_2017, devlin_bert_2019,conneau_cross-lingual_2019}. It consists of an embedding layer and $L$ stacked transformer blocks with self-attention modules. Each input token $x_i$ will be encoded into a vector space as the initial hidden vector $h_i^0$. Then, in each transformer block, the hidden vector of the $i$-th token $h_i^l$ is computed from the self-attentive fusion of all hidden vectors output from the previous layer:
\begin{equation}
    h^l=(h_1^l,h_2^l,\cdots,h_S^l)=f^l(h^{l-1}).
\end{equation}
We finally get the contextualized token representation $f(x)=f^L(f^{L-1}(\cdots f^1(h^0)))$.

\paragraph{Cross-lingual Pre-training}
Masked language modeling (MLM)~\citep{devlin_bert_2019} and Translation language modeling (TLM)~\citep{conneau_cross-lingual_2019} are two typical tasks for cross-lingual pre-training. MLM is conducted on monolingual corpora. A randomly selected subset of input tokens will be replaced by a special [MASK] token or another random token, and models learn to recover these corrupted tokens according to the context. TLM extends MLM to cross-lingual scenarios with the following objective:
\begin{equation}
    \mathcal{L}_{TLM}(x,y)=\ell\left(x\oplus y, f(m(x)\oplus m(y))\right),
\end{equation}
where $\oplus$ denotes sequence concatenation operator and $m$ denotes element-wise random replacement. During training, models can predict the masked token using the unmasked token in the translation. In this way, models learn cross-lingual token-level alignment using the parallel corpora.

However, TLM is designed for a cross-encoder architecture in which tokens from the source and target sentences are mutually accessible in intermediate layers. As a result, models trained with TLM may rely on this information exchange, which is not available during the inference stage when sentences are independently encoded. Additionally, computing TLM requires an extra feedforward propagation, which inputs concatenated sentence pairs, resulting in increased training costs. Our proposed representation translation learning task can overcome both the weaknesses.

\subsection{Model Structure}
Our dual-alignment pre-training framework contains two transformer models: dual encoder model $f$ and representation translation learning (RTL) head $g$.

For the encoder model, we adopt the most popular BERT architecture with 12 layers of transformer encoder blocks, 12 attention heads, and 768-dimension hidden states. Following~\citet{devlin_bert_2019}, we prepend a special token [CLS] to the input:

\begin{equation}
    f(x)=f({\rm [CLS]},x_1,\dots,x_S).
\end{equation}

We take the hidden vector of CLS token $h_{cls}^L$ as the representation of the whole sentence $f_s(x)$. Like other multilingual language models, our model is language-agnostic, which means all languages share the same single transformer. 

The RTL head is a stack of $K$ transformer encoder blocks with a vocabulary prediction head at the top. The function of RTL head is to reconstruct the translation sentence $y$ from the token representations of the source sentence $h^L$ (source sentences indicate non-English sentences in this paper):

\begin{equation}
    \begin{split}
        &g(h,y)=\pi\left(W^Tg^K\left(g^{K-1}\left(\cdots g^0(h,y)\right)\right)\right),\\
        &g^0(h,y)\!=\!(h_1^L\!,\!\cdots,\!h_{S_x}^L,\!\underbrace{{\rm [MASK]},\!\cdots\!,\!{\rm [MASK]}}_{\times S_y}),
    \end{split}
\end{equation}
where $\pi$ is softmax function and $W$ is the weight matrix of the vocabulary prediction head. In our experiments, we find a small RTL head with $K=2$ performs best generally.

\subsection{Pre-training Tasks}\label{sec:framework}
To achieve both sentence-level and token-level alignment, we design a pre-training framework consisting of two tasks: translation ranking task and representation translation learning task. These two objectives are leveraged simultaneously during training. The whole procedure is depicted in Figure~\ref{fig:arch}.
\subsubsection{Translation Ranking}
Dual encoder models trained with the translation ranking (TR) task have been proven effective in learning cross-lingual embeddings~\citep{yang_improving_2019, labse, chi-etal-2021-infoxlm}. These models learn to maximize the similarity of the embedding pairs of parallel sentences and the dissimilarity of mismatched pairs. Therefore, they are well suited for solving retrieval and mining tasks that use inner product as ranking metrics. Following~\citep{labse}, we formulate the training task as follows:
\begin{equation}
    \mathcal{L}_{TR}=-\frac{1}{N}\sum_{i=1}^N\log\frac{e^{\phi(x_i,y_i)}}{\sum_{j=1}^Be^{\phi(x_i,y_j)}},
\end{equation}
where $B$ is the batch size and $\phi(x, y)$ is defined as the similarity of the representation of each text, typically $f_s(x)^Tf_s(y)$. In this paper, we use the hidden vector of CLS token to represent the sentence.
\subsubsection{Representation Translation Learning}
Minimizing $\mathcal{L}_{TR}$ essentially maximize the lower bound of the mutual information $I(x;y)$~\citep{oord_representation_2018,chi-etal-2021-infoxlm}. However, it is hard for models to find an embedding perfectly containing all information of the sentence. Consequently, models may only pay attention to the high-level global information and neglect some local token-level information. To this end, we add an auxiliary loss to force the models to preserve the token-level information throughout the entire model:
\begin{equation}
    \mathcal{L}_{RTL}=\frac{1}{S}\sum_{i=1}^S{CE}(g(f_*(x),y)_i,y_i),
\end{equation}
where $f_*(x)$ denotes all hidden vectors of $x$ except CLS and CE denotes cross entropy. It is worth noting that we do not involve the CLS token in calculating RTL objective because we find it will make translation ranking objective hard to converge. To train the RTL head with a stable and consistent target, the reconstruction direction is always from non-English sentences to their English translations.

Combining with the translation ranking objective we get the final loss:
\begin{equation}
    \mathcal{L}_{DAP}=\mathcal{L}_{TR}+\mathcal{L}_{RTL}.
\end{equation}
As RTL does not need an extra feedforward propagation, RTL only introduces a little computation and will not slow down the pre-training significantly. The only time-consuming operation is the softmax over the huge vocabulary which can be further relieved by techniques like negative sampling and hierarchical softmax (not used in our experiments).

\section{Experiments}
In this section, we first describe the training setup. Then we compare our method with previous works on three sentence-level cross-lingual tasks.
\subsection{Pre-training data}
Following~\citet{laser} we collect parallel training data for 36 languages (used in XTREME Tatoeba benchmark) by combining Europarl, United Nations Parallel Corpus, OpenSubtitles, Tanzil, CCMatrix and WikiMatrix corpora, which are downloaded from OPUS website~\citep{opus}. As stated in section~\ref{sec:framework}, we align all other languages with English, so we only collect parallel corpora that contain English. For each non-English language, we retain at most 1 million sentence pairs at random. The whole dataset has 5.7GB data, which is far less than typical large-scale pre-training~\citep{labse, chi-etal-2021-infoxlm}, but our method still achieves performance comparable with the state-of-the-art.

\begin{table*}[t]
    \centering
    \begin{tabular}{l|ccc|ccc}
    \toprule
         Direction& \multicolumn{3}{c}{xx$\to$en} &\multicolumn{3}{c}{en$\to$xx} \\
         \hline
         Model & 14 langs & 28 langs & 36 langs & 14 langs & 28 langs & 36 langs \\
         \hline
         InfoXLM & 77.8 & - & -  & 80.6 & - & -  \\
         LaBSE&-&-&-&-&-&\underline{93.7}\\
         \hline
         mBERT$^*$ & - & -& - &45.6 & 45.1 & 38.7\\
         mBERT (recomputed) & 42.5 & 42.2& 36.9 &43.8 & 43.3 & 37.2\\
         mBERT+TR & 94.0 & 93.8 & 90.1 & 93.2 & 93.4 & 90.1\\
         mBERT+TR+TLM & 94.1 &  93.8 & 90.2 & 93.5 & 93.5 & 90.3\\
         \textbf{mBERT+DAP}&\textbf{94.7} & \textbf{94.7} & \textbf{90.9} & \textbf{94.2} & \textbf{94.6} & \textbf{91.2} \\
         \hline
         XLM-R$^*$&-&-&-& 60.6&63.7 &57.7\\
         XLM-R (recomputed)&59.4&60.1&55.3& 57.5&58.9 &53.3\\
         XLM-R+TR& 93.8 & 94.2 & \textbf{91.6} & 91.2 & 91.2 & 86.4\\
         XLM-R+TR+TLM& 93.2 & 92.8 & 89.2 & 94.4 & 94.5 & 92.4\\
         \textbf{XLM-R+DAP}& \textbf{95.0} & \textbf{94.7} & 91.3 &\textbf{95.1} & \textbf{95.2} & \textbf{92.7}\\
    \bottomrule
    \end{tabular}
    \caption{Average accuracy on Tatoeba bitext retrieval task. Direction "xx$\to$en" means retrieval is performed over the English corpora, and vice versa. 14 langs and 28 langs mean different subsets of all 36 languages. For mBERT and XLM-R models, we report both the best implementation before (Results with * are taken from~\citep{hu_xtreme_2020}) and our recomputed accuracy. Results of InfoXLM and LaBSE are taken from their papers.  For LaBSE we take the result using mBERT vocabulary for fair comparison. Bold font means that model performs the best among its group. We use underline to identify a state-of-the-art method that outperforms all our variants.}
    \label{fig:tatoeba}
\end{table*}

\subsection{Implementation Details}
We initialize the encoder model from multilingual BERT base or XLM-R base, respectively, using the checkpoint published on Huggingface model hub, and initialize the $K$-layer RTL head from the last $K$ transformer layers by the corresponding encoder model. The maximum sentence length is restricted to 32 tokens, and sentences longer than 32 tokens will be truncated. We train the model for 100,000 steps using the AdamW optimizer with a learning rate of 5e-5 and a total batch size of 1024 on 8 Tesla V100 GPUs for 1 day. The results reported are the average of three different seeds.

\subsection{Compared models}
To demonstrate the effectiveness of our proposed Representation Translation Learning, we first compare it with the base models (mBERT or XLM-R) and their TR-finetuned versions. Additionally, we also introduce a variant of our method that leverages TLM.

Furthermore, we also compare our approach with two state-of-the-art multilingual language models, InfoXLM~\citep{chi-etal-2021-infoxlm} and LaBSE~\citep{labse}. It is worth noting that InfoXLM and LaBSE use 10 times more training data than our method and are trained longer with a larger batch size.
\setlength\tabcolsep{3.5pt}
\begin{table*}[t]
    \centering
    \begin{tabular}{l|ccc|ccc|ccc|ccc|c}
        \toprule
        \multirow{2}{*}{Model}&\multicolumn{3}{c|}{fr-en}&\multicolumn{3}{c|}{de-en}&\multicolumn{3}{c|}{ru-en}&\multicolumn{3}{c|}{zh-en}&Avg\\
        &P&R&F&P&R&F&P&R&F&P&R&F&F\\
        \hline
        LaBSE & 96.3 & 93.6 & 95.0 & 99.4 & 95.4 & 97.3 & 99.3 & 93.1 & 96.1 & 90.4 & 88.3 & 89.4 & 94.5 \\
        \hline
         mBERT (recomputed) & 75.1 & 68.2 & 71.5 & 77.8 & 69.0 & 73.1 & 70.1 & 52.9 & 60.3 & 63.1 & 50.6 & 56.2 & 65.3\\
         mBERT+TR & 96.1 & 90.9 & 93.4 & 98.8 & 94.0 & 96.3 & 98.4 & 89.8 & 93.9 & 96.0 & 93.8 & 94.9 & 94.6\\
         mBERT+TR+TLM & 95.6 & 90.9 & 93.2 & 98.3 & 94.0 & 96.1 & 97.0 & 89.7 & 93.2 & 93.9 & 95.7 & 94.8 & 94.3\\
         \textbf{mBERT+DAP}&95.1 & 94.1 & 94.6 & 98.1 & 94.7 & 96.4 & 98.6 & 91.4 & 94.9 & 95.7 & 94.2 & 94.9 & \textbf{95.2} \\
         \hline
         XLM-R (recomputed)&  81.3 & 68.2 & 74.2 & 86.6 & 77.0 & 81.5 & 87.6 & 74.0 & 80.2 & 77.0 & 54.9 & 64.1 & 75.0\\
         XLM-R+TR& 92.6 & 92.1 & 92.4 & 96.3 & 94.6 & 95.4 & 97.3 & 91.0 & 94.0 & 96.6 & 87.5 & 91.8 & 93.4\\
         XLM-R+TR+TLM& 91.4 & 91.6 & 91.5 & 94.0 & 95.5 & 94.7 & 94.4 & 90.9 & 92.7 & 92.8 & 90.3 & 91.5 & 92.6\\
         \textbf{XLM-R+DAP}& 95.3 & 93.1 & 94.2 & 99.0 & 95.2 & 97.1 & 98.1 & 93.3 & 95.6 & 96.7 & 92.6 & 94.6 & \textbf{95.4}\\
        \bottomrule
    \end{tabular}
    \caption{Evaluation on BUCC training set. The thresholds are chosen to achieve the optimal F1 score. }
    \label{tab:bucc_train}
\end{table*}
\begin{table*}[t]
    \centering
    \begin{tabular}{l|ccc|ccc|ccc|ccc|c}
        \toprule
        \multirow{2}{*}{Model}&\multicolumn{3}{c|}{fr-en}&\multicolumn{3}{c|}{de-en}&\multicolumn{3}{c|}{ru-en}&\multicolumn{3}{c|}{zh-en}&Avg\\
        &P&R&F&P&R&F&P&R&F&P&R&F&F\\
        \hline
        LaBSE & 92.8 & 82.5 & 87.4 & 96.6 & 85.2 & 90.5 & 91.2 & 85.9 & 88.5 & 85.5 & 70.4 & 77.2 & 85.9 \\
        \hline
         mBERT$^*$ & - & - & 62.6 & - & - & 62.5 & - & - & 51.8 & - & - & 50.0 & 56.7\\
         mBERT (recomputed) & 80.1 & 42.1 & 55.2 & 83.7 & 38.2 & 52.5 & 69.1 & 28.9 & 40.8 & 65.8 & 20.2 & 30.9 & 44.8\\
         mBERT+TR & 93.6 & 75.2 & 83.4 & 97.3 & 77.1 & 86.0 & 91.3 & 77.2 & 83.6 & 93.0 & 69.7 & 79.7 & 83.2\\
         mBERT+TR+TLM & 92.4 & 75.0 & 82.8 & 96.2 & 78.2 & 86.3 & 90.1 & 77.2 & 83.1 & 90.9 & 75.8 & 82.6 & 83.7\\
         \textbf{mBERT+DAP}&92.1 & 83.4 & 87.6 & 96.2 & 83.6 & 89.5 & 90.1 & 82.4 & 86.1 & 92.5 & 75.7 & 83.3 & \textbf{86.6} \\
         \hline
         XLM-R$^*$&  - & - & 67.5 & - & - & 66.5 & - & - & 73.5 & - & - & 56.7 & 66.0\\
         XLM-R (recomputed)& 85.9 & 47.3 & 61.0 & 88.6 & 48.3 & 62.5 & 85.8 & 54.3 & 66.5 & 77.7 & 27.3 & 40.4 & 57.6\\
         XLM-R+TR& 89.7 & 79.1 & 84.1 & 94.2 & 80.3 & 86.7 & 89.6 & 80.2 & 84.7 & 92.2 & 66.1 & 77.0 & 83.1\\
         XLM-R+TR+TLM& 88.1 & 75.8 & 81.5 & 91.2 & 79.8 & 85.1 & 86.3 & 80.6 & 83.4 & 89.6 & 72.6 & 80.2 & 82.5\\
         \textbf{XLM-R+DAP}& 92.1 & 82.1 & 86.8 & 96.6 & 81.1 & 88.2 & 89.5 & 88.1 & 88.8 & 93.7 & 75.0 & 83.3 & \textbf{86.8}\\
        \bottomrule
    \end{tabular}
    \caption{Evaluation on BUCC test set. The thresholds are chosen to achieve the optimal F1 score on training set. For mBERT and XLM-R models, we report both the best implementation before (Results with * are taken from~\citep{hu_xtreme_2020}) and our recomputed scores.}
    \label{tab:bucc_test}
\end{table*}
\subsection{Bitext Retrieval}
In bitext retrieval, given a query sentence from source language, models need to retrieve the most relevant sentence among a collection of sentences in the target language. Following previous works~\citep{labse,chi-etal-2021-infoxlm,laser}, we use the Tatoeba dataset to evaluate our pre-training framework in a zero-shot manner.

Tatoeba contains parallel sentences in more than 300 languages, and we use the 36 languages version from XTREME benchmark~\citep{hu_xtreme_2020}. Each language has up to 1000 sentences paired with English.
\paragraph{Results}
 We test on all 36 languages and report the average accuracy over 14 languages tested in LASER~\citep{laser} and 36 languages tested in XTREME. Besides, we set up a new group of 28 languages based on our observation of the low-resource test languages. Among the original 36 languages, some scarce languages have less than 1000 sentence pairs, and some of them even only have about 200 sentence pairs, and we observe that the accuracy of these languages is inconsistent between the two retrieval directions ("en$\to$xx" and "xx$\to$en" with a difference more than 30\%) and also significantly lower than other languages with abundant resources. This indicates that the results obtained from small test sets are not as reliable as those from larger test sets. Therefore, we report a 28-language version where all languages contain 1000 test pairs. The retrieval accuracy for each language is reported in the appendix~\ref{app:tatoeba}.
 
In Table~\ref{fig:tatoeba}, we observe that our DAP method outperforms all other variants significantly. mBERT and XLM-R perform the worst because they lack a sentence-level objective. TLM improves TR's performance in the direction "en$\to$xx" but hurts direction "xx$\to$en". By contrast, DAP brings consistent improvement. Compared with the two state-of-the-art methods, our method performs much better than InfoXLM and only slightly falls behind LaBSE. Considering the training cost, we think this result has demonstrated DAP's potential.
\subsection{Bitext Mining}
In bitext mining, models need to detect the parallel sentence pairs (e.g., translations) from a pair of monolingual corpus. We use the BUCC 2018 dataset~\citep{bucc} to perform evaluations, which contains four language pairs: fr-en, de-en, ru-en and zh-en. Each corpus contains 150k to 1.2M unpaired sentences and gold labels telling which sentences are translation pairs.

Following~\citet{artetxe_margin-based_2019}, we employ the ratio between the cosine of a given candidate and the average cosine of its neighbours in both directions. The training set is used to learn the best threshold~\citep{schwenk_filtering_2018} to decide which pairs should be selected. More details of the scoring function and threshold can be found in appendix~\ref{app:bucc}.
\setlength\tabcolsep{1.8pt}
\begin{table*}
    \centering
    \begin{tabular}{l|cccccccccccccccc}
         \toprule
         Model & en&fr&es&de&el&bg&ru&tr&ar&vi&th&zh&hi&sw &ur&Avg\\
         \hline
         InfoXLM & 86.4 & 80.3 & 80.9 & 79.3 & 77.8 & 79.3 & 77.6 & 75.6 & 74.2 & 77.1 & 74.6 & 77.0 & 72.2 & 67.5 & 67.3 & \underline{76.5}\\
         LaBSE & 85.4 & 80.2 & 80.5 & 78.8 & 78.6 & 80.1 & 77.5 & 75.1 & 75.0 & 76.5 & 69.0 & 75.8 & 71.9 & 71.5 & 68.1 & \underline{76.3}\\
         \hline
         mBERT & 82.1 & 74.4 & 74.9 & 71.2 & 67.9 & 69.5 & 69.6 & 62.8 & 66.2 & 70.6 & 54.6 & 69.7 & 60.4 & 50.9 & 58.0 & 66.8\\
         mBERT+TR & 82.0 & 74.3 & 75.1 & 72.9 & 69.9 & 73.1 & 70.6 & 68.6 & 67.4 & 73.6 & 61.3 & 70.8 & 65.0 & 62.6 & 61.0 & 69.9\\
         mBERT+TR+TLM & 82.8 & 75.2 & 74.4 & 72.0 & 69.3 & 70.6 & 69.4 & 66.1 & 66.1 & 70.6 & 58.9 & 67.3 & 63.7 & 60.6 & 59.5 & 68.4\\
         \textbf{mBERT+DAP} & 81.8 & 75.6 & 76.2 & 74.4 & 72.6 & 74.9 & 72.0 & 71.3 & 69.7 & 74.4 & 63.6 & 72.3 & 67.3 & 67.3 & 63.2 & \textbf{71.8}\\
         \hline
         XLM-R & 83.8 & 77.6 & 78.2 & 75.4 & 75.0 & 77.0 & 74.8 & 72.7 & 72.0 & 74.5 & 72.1 & 72.9 & 69.6 & 64.2 & 66.0 & 73.7\\
         XLM-R+TR & 83.5 & 76.4 & 76.8 & 75.7 & 74.2 & 76.2 & 74.6 & 71.8 & 71.1 & 74.2 & 69.1 & 72.9 & 68.8 & 66.8 & 65.2 & 73.1\\
         XLM-R+TR+TLM & 84.6 & 77.4 & 76.9 & 74.9 & 68.1 & 69.8 & 69.4 & 68.1 & 61.7 & 68.9 & 62.6 & 66.9 & 61.4 & 61.7 & 57.5 & 68.7\\
         \textbf{XLM-R+DAP}& 82.9 & 77.0 & 77.7 & 75.7 & 75.2 & 76.0 & 74.7 & 73.1 & 72.5 & 74.2 & 71.9 & 73.0 & 69.8 & 70.5 & 66.0 & \textbf{74.0}\\
         \bottomrule
    \end{tabular}
    \caption{Accuracy for XNLI cross-lingual natural language inference. Results of InfoXLM are taken from their paper.}
    \label{tab:xnli}
\end{table*}
\paragraph{Results}
Table~\ref{tab:bucc_train} shows the precision, recall and F1 score for four language pairs on training set after optimization. The results of LaBSE are produced using the checkpoints publicized in Huggingface model hub. We do not report the results of InfoXLM because this task was not evaluated in the original paper and we failed to produce reasonable results.

Our method outperforms all variants and even LaBSE, which means our model learns an embedding space with better separability. When testing the optimized model on test set, our model shows remarkable generalization ability and enlarges the gap against other methods as shown in Table~\ref{tab:bucc_test}. We outperform the state-of-the-art LaBSE by 0.9\% and other variants by at least 3.0\%. Similar to the retrieval task, mBERT and XLM-R perform the worst. TLM brings improvements for zh-en but gets worse for fr-en. DAP consistently performs the best on all metrics. Furthermore, the improvement observed in DAP's performance is larger in comparison to the retrieval task. This indicates that DAP is more effective in enhancing performance on complex tasks, suggesting its potential as a valuable tool for addressing challenging problems.
\subsection{Cross-lingual Natural Language Inference}
Natural language inference (NLI) is a well-known task to evaluate models' classification performance under fine-tuning. The goal is to predict the relationship between the input sentence pair. The candidate relationships are entailment, contradiction and neutral. XNLI~\citep{conneau_xnli_2018} extends NLI to the multilingual setting of 15 languages. Following~\citet{chi-etal-2021-infoxlm}, we fine-tune the model with the English training set and directly evaluate on test sets of other languages. The hyperparameters of fine-tuning are reported in the appendix~\ref{app:xnli}.
\paragraph{Results}
Table~\ref{tab:xnli} shows accuracy for 15 languages. We observe that the differences between variants are relatively small compared with retrieval and mining tasks. We think this is because judging the relationship between two sentences does not rely on cosine similarity, so the pre-training cannot be directly transferred to the downstream task. mBERT variants all show positive results and DAP has the largest improvement. But for XLM-R variants, only DAP maintains the performance as the base model. The TR and TLM variants suffer from performance degradation. We think this is because XLM-R has already been a well-trained multilingual model and our continued pre-training is insufficient to improve the classification capacity. However, we demonstrate DAP will not harm classification performance for a well-trained base model.

\setlength\tabcolsep{4pt}
\begin{table}[t]
    \centering
    \begin{tabular}{c|ccc}
        \toprule
         Direction& Tatoeba & BUCC & XNLI \\
         \hline
         xx$\to$en& \textbf{91.0} & \textbf{86.6} & \textbf{71.8} \\
         en$\to$xx& 90.5 & 84.1 & 69.3\\
         Both& 90.8 & 86.3 & 70.5\\
         \bottomrule
    \end{tabular}
    \caption{Performance of different RTL directions across three tasks. "xx$\to$en" means RTL head reconstructs English sentences using non-English token representations, and vice versa. "Both" means we calculate the RTL loss from both directions on half of the batch respectively and take the average.}
    \label{tab:direction}
\end{table}

\section{Analysis}
In this section, we conduct experiments to get a deeper understanding of DAP. In each setting, we report the average accuracy over 36 languages and two retrieval directions on Tatoeba, average F1 score on BUCC test set and average accuracy on XNLI. All variants are trained from mBERT.
\subsection{Translation Direction}
In our method, the RTL head only learns to translate from non-English to English. Here we investigate if the opposite direction can help the pre-training. To remind the model of the language to be reconstructed, we add language embeddings to the representation before the RTL head like TLM.

As shown in Table~\ref{tab:direction}, translating from English to non-English performs much worse than the opposite direction. Also, the mixed-up training gets an intermediate performance. We attribute the difference between the two directions to the dispersion of the objective. We assume that RTL aligns the source language's representation towards the target language. So, if the reconstruction target keeps switching among different languages, it will make RTL hard to converge.
\begin{figure}[t]
    \centering
    \includegraphics[width=0.99\linewidth]{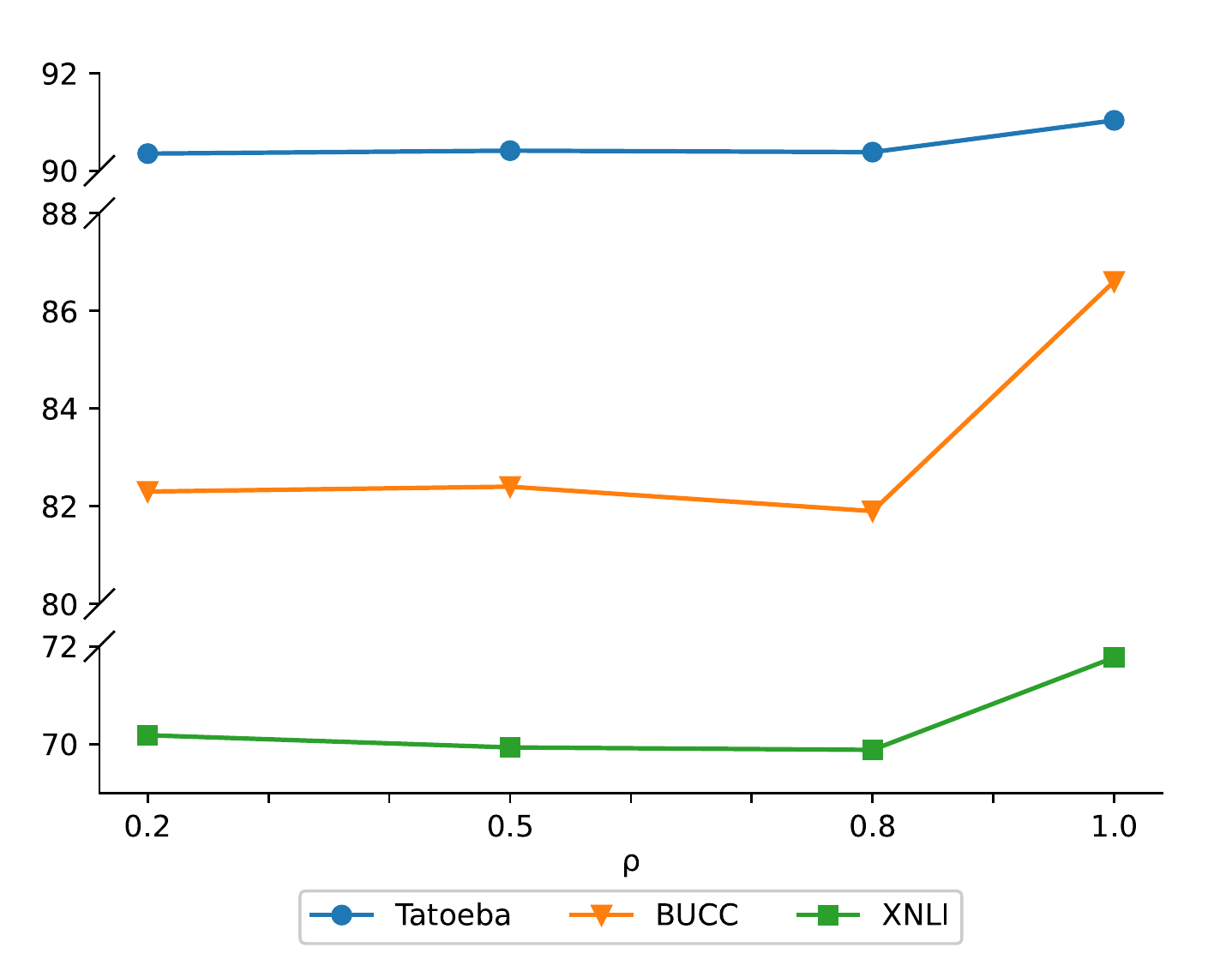}
    \caption{Performance of varying reconstruction ratios across three tasks.}
    \label{fig:reconstruction}
\end{figure}
\subsection{Reconstruction Ratio}
To better understand the objective of the RTL task, we conduct experiments where RTL head only needs to reconstruct partial target sentences with the other target token representations accessible. The tokens to reconstruct are selected randomly with probability $\rho$. Larger $\rho$ will make the RTL task harder.

From Figure~\ref{fig:reconstruction}, we can find the variants with $\rho<1$ have similar performance on all tasks and there is a steep increase at $\rho=1$. We think this is because the unmasked target token representations cause information leakage, so the RTL head does not need to learn the alignment from source sentences.

\subsection{Complexity of RTL head}
We investigate the relation between the RTL head's complexity and the pre-training performance. We set $K=1,2,3,4$ to give RTL head different capabilities to extract aligned information from the representation of the source sentence.

\begin{figure}[t]
    \centering
    \includegraphics[width=0.98\linewidth]{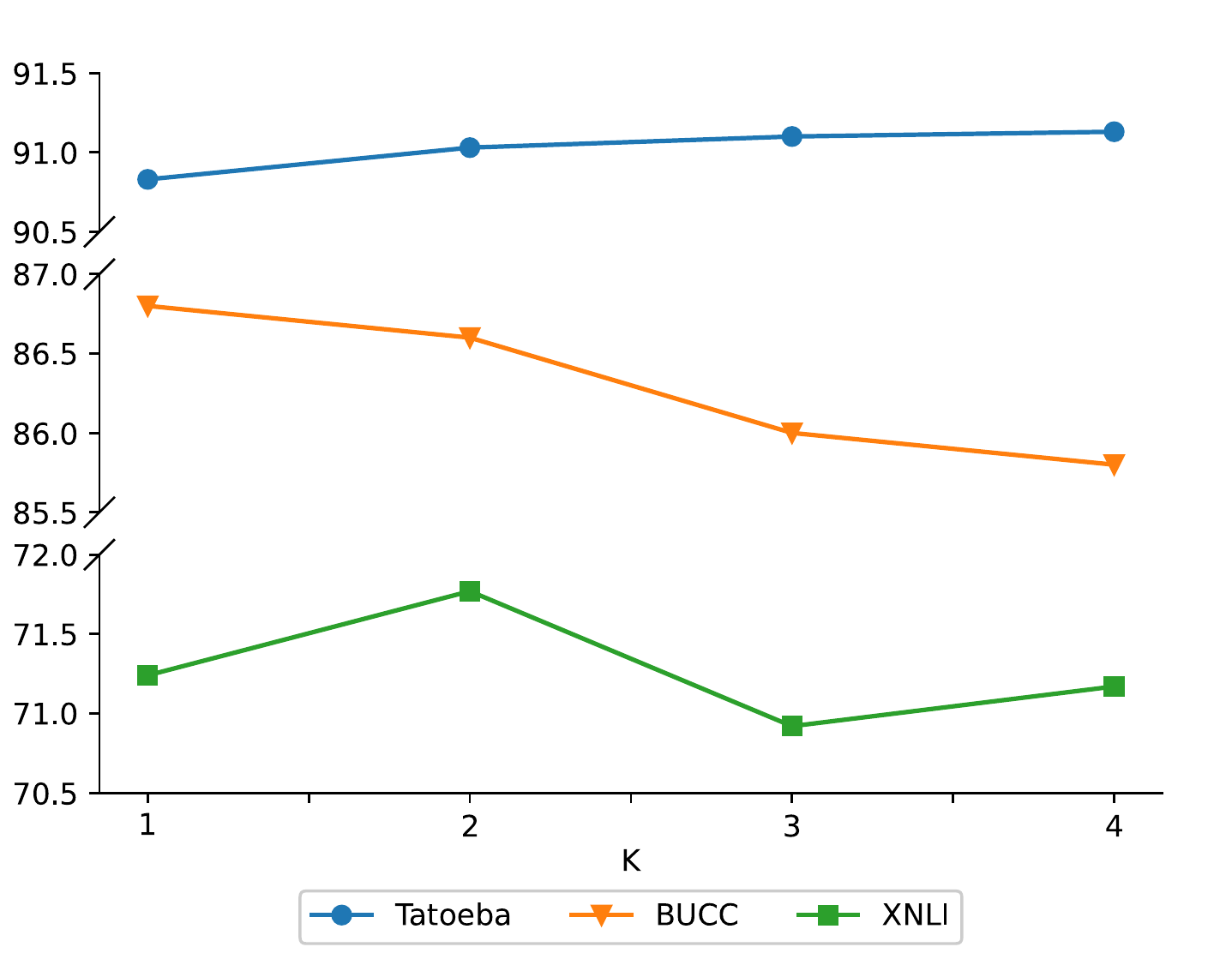}
    \caption{Performance of varying numbers of RTL head layers across three tasks.}
    \label{fig:complexity}
\end{figure}

In Figure~\ref{fig:complexity}, the three tasks show different tendencies with regard to RTL head's complexity. Only the accuracy on Tatoeba keeps increasing along with $K$ but the gain from larger $K$ is declining especially after $K=2$. For the other two tasks, larger $K$ brings a negative effect. We hypothesize that a smaller $K$ that makes RTL task harder will enforce the model to generate more informative representations. Setting $K=2$ achieves the best general cross-lingual performance across three tasks. 

\subsection{Computational Efficiency}
Computational efficiency is an important factor when designing pre-training tasks. A more efficient method enables models to train on a larger dataset for more steps. We calculate the feedforward floating point operations (FLOPs) for our method and TLM, respectively. In addition, we report the training latency in our training environment. We measure the latency with a total batch size of 512 on 8 Tesla V100 GPUs using PyTorch distributed data parallel.
\begin{table}[t]
    \centering
    \begin{tabular}{l|cc}
    \toprule
         Model & FLOPs & Latency  \\
         \hline
         mBERT+TR & 11.0G & 0.51 \\
         mBERT+TR+TLM & 33.7G & 1.34 \\
         \textbf{mBERT+DAP} & 16.5G & 0.88 \\
    \bottomrule
    \end{tabular}
    \caption{Computational efficiency of different pre-training methods. The unit of latency is milliseconds per sample.}
    \label{tab:efficiency}
\end{table}

From Table~\ref{tab:efficiency}, we can find DAP only increases the training cost by about 50\% against the TR-only baseline, which can be further improved if we use negative sampling to reduce the softmax over the huge vocabulary. By contrast, TLM introduces a training cost of more than 150\% due to the extra feedforward propagation through the 12-layer encoder. Therefore, DAP is more efficient and scalable for cross-lingual pre-training.

\section{Conclusion}
In this paper, we find that token-level alignment is crucial for cross-lingual tasks. Based on this observation, we present a dual-alignment pre-training framework for cross-lingual sentence embedding that enables both sentence-level and token-level alignment. The framework consists of a translation ranking task and a newly proposed representation translation learning task, which encourages the token representation to contain all information from its translation counterpart in an efficient way.

We train our models on a moderate-size corpus. The model trained with DAP significantly outperforms variants without token-level alignment or using TLM as the alignment task across three sentence-level cross-lingual tasks, and achieves performance comparable with those state-of-the-art pre-training work trained on 10 times more data with larger batch size and training steps. These results show our approach brings essential improvement for cross-lingual sentence embedding.

\section*{Limitations}
Although our method is efficient and scalable, we have not conducted pre-training on large-scale corpora due to limited computational resources. The quality and quantity of data are crucial factors for a pre-training model. As our model only covers 36 languages, it cannot provide services for many rare languages. This paper just proposes a new pre-training direction and does not use many training tricks. Exploring DAP's full capability is left for future work.

Besides, RTL task is not the only possible token-alignment task for our DAP framework. Other objectives based on token representations are also worth investigating. The best objective form is still under research.

\bibliography{anthology,custom}

\begin{thebibliography}{20}
\expandafter\ifx\csname natexlab\endcsname\relax\def\natexlab#1{#1}\fi

\bibitem[{Artetxe and Schwenk(2019{\natexlab{a}})}]{artetxe_margin-based_2019}
Mikel Artetxe and Holger Schwenk. 2019{\natexlab{a}}.
\newblock \href {https://doi.org/10.18653/v1/p19-1309} {Margin-based {Parallel}
  {Corpus} {Mining} with {Multilingual} {Sentence} {Embeddings}}.
\newblock In \emph{Proceedings of the 57th {Conference} of the {Association}
  for {Computational} {Linguistics}, {ACL} 2019, {Florence}, {Italy}, {July}
  28- {August} 2, 2019, {Volume} 1: {Long} {Papers}}, pages 3197--3203.
  Association for Computational Linguistics.

\bibitem[{Artetxe and Schwenk(2019{\natexlab{b}})}]{laser}
Mikel Artetxe and Holger Schwenk. 2019{\natexlab{b}}.
\newblock \href {https://doi.org/10.1162/tacl_a_00288} {Massively
  {Multilingual} {Sentence} {Embeddings} for {Zero}-{Shot} {Cross}-{Lingual}
  {Transfer} and {Beyond}}.
\newblock \emph{Transactions of the Association for Computational Linguistics},
  7:597--610.

\bibitem[{Chi et~al.(2021)Chi, Dong, Wei, Yang, Singhal, Wang, Song, Mao,
  Huang, and Zhou}]{chi-etal-2021-infoxlm}
Zewen Chi, Li~Dong, Furu Wei, Nan Yang, Saksham Singhal, Wenhui Wang, Xia Song,
  Xian-Ling Mao, Heyan Huang, and Ming Zhou. 2021.
\newblock \href {https://doi.org/10.18653/v1/2021.naacl-main.280} {{I}nfo{XLM}:
  An information-theoretic framework for cross-lingual language model
  pre-training}.
\newblock In \emph{Proceedings of the 2021 Conference of the North American
  Chapter of the Association for Computational Linguistics: Human Language
  Technologies}, pages 3576--3588, Online. Association for Computational
  Linguistics.

\bibitem[{Chidambaram et~al.(2019)Chidambaram, Yang, Cer, Yuan, Sung, Strope,
  and Kurzweil}]{chidambaram_learning_2019}
Muthu Chidambaram, Yinfei Yang, Daniel Cer, Steve Yuan, Yunhsuan Sung, Brian
  Strope, and Ray Kurzweil. 2019.
\newblock \href {https://doi.org/10.18653/v1/W19-4330} {Learning
  {Cross}-{Lingual} {Sentence} {Representations} via a {Multi}-task
  {Dual}-{Encoder} {Model}}.
\newblock In \emph{Proceedings of the 4th {Workshop} on {Representation}
  {Learning} for {NLP} ({RepL4NLP}-2019)}, pages 250--259, Florence, Italy.
  Association for Computational Linguistics.

\bibitem[{Conneau et~al.(2020{\natexlab{a}})Conneau, Khandelwal, Goyal,
  Chaudhary, Wenzek, Guzmán, Grave, Ott, Zettlemoyer, and
  Stoyanov}]{conneau_unsupervised_2020}
Alexis Conneau, Kartikay Khandelwal, Naman Goyal, Vishrav Chaudhary, Guillaume
  Wenzek, Francisco Guzmán, Edouard Grave, Myle Ott, Luke Zettlemoyer, and
  Veselin Stoyanov. 2020{\natexlab{a}}.
\newblock \href {https://doi.org/10.18653/v1/2020.acl-main.747} {Unsupervised
  {Cross}-lingual {Representation} {Learning} at {Scale}}.
\newblock In \emph{Proceedings of the 58th {Annual} {Meeting} of the
  {Association} for {Computational} {Linguistics}}, pages 8440--8451, Online.
  Association for Computational Linguistics.

\bibitem[{Conneau and Lample(2019)}]{conneau_cross-lingual_2019}
Alexis Conneau and Guillaume Lample. 2019.
\newblock \href
  {https://proceedings.neurips.cc/paper/2019/hash/c04c19c2c2474dbf5f7ac4372c5b9af1-Abstract.html}
  {Cross-lingual {Language} {Model} {Pretraining}}.
\newblock In \emph{Advances in {Neural} {Information} {Processing} {Systems}},
  volume~32. Curran Associates, Inc.

\bibitem[{Conneau et~al.(2018)Conneau, Rinott, Lample, Williams, Bowman,
  Schwenk, and Stoyanov}]{conneau_xnli_2018}
Alexis Conneau, Ruty Rinott, Guillaume Lample, Adina Williams, Samuel Bowman,
  Holger Schwenk, and Veselin Stoyanov. 2018.
\newblock \href {https://doi.org/10.18653/v1/D18-1269} {{XNLI}: {Evaluating}
  {Cross}-lingual {Sentence} {Representations}}.
\newblock In \emph{Proceedings of the 2018 {Conference} on {Empirical}
  {Methods} in {Natural} {Language} {Processing}}, pages 2475--2485, Brussels,
  Belgium. Association for Computational Linguistics.

\bibitem[{Conneau et~al.(2020{\natexlab{b}})Conneau, Wu, Li, Zettlemoyer, and
  Stoyanov}]{conneau_emerging_2020}
Alexis Conneau, Shijie Wu, Haoran Li, Luke Zettlemoyer, and Veselin Stoyanov.
  2020{\natexlab{b}}.
\newblock \href {https://doi.org/10.18653/v1/2020.acl-main.536} {Emerging
  {Cross}-lingual {Structure} in {Pretrained} {Language} {Models}}.
\newblock In \emph{Proceedings of the 58th {Annual} {Meeting} of the
  {Association} for {Computational} {Linguistics}}, pages 6022--6034, Online.
  Association for Computational Linguistics.

\bibitem[{Devlin et~al.(2019)Devlin, Chang, Lee, and
  Toutanova}]{devlin_bert_2019}
Jacob Devlin, Ming-Wei Chang, Kenton Lee, and Kristina Toutanova. 2019.
\newblock \href {https://doi.org/10.18653/v1/N19-1423} {{BERT}: {Pre}-training
  of {Deep} {Bidirectional} {Transformers} for {Language} {Understanding}}.
\newblock In \emph{Proceedings of the 2019 {Conference} of the {North}
  {American} {Chapter} of the {Association} for {Computational} {Linguistics}:
  {Human} {Language} {Technologies}, {Volume} 1 ({Long} and {Short} {Papers})},
  pages 4171--4186, Minneapolis, Minnesota. Association for Computational
  Linguistics.

\bibitem[{Feng et~al.(2022)Feng, Yang, Cer, Arivazhagan, and Wang}]{labse}
Fangxiaoyu Feng, Yinfei Yang, Daniel Cer, Naveen Arivazhagan, and Wei Wang.
  2022.
\newblock \href {https://doi.org/10.18653/v1/2022.acl-long.62}
  {Language-agnostic {BERT} {Sentence} {Embedding}.}
\newblock In \emph{Proceedings of the 60th {Annual} {Meeting} of the
  {Association} for {Computational} {Linguistics} ({Volume} 1: {Long}
  {Papers}), {ACL} 2022, {Dublin}, {Ireland}, {May} 22-27, 2022}, pages
  878--891.

\bibitem[{Guo et~al.(2018)Guo, Shen, Yang, Ge, Cer, Ábrego, Stevens, Constant,
  Sung, Strope, and Kurzweil}]{guo_effective_2018}
Mandy Guo, Qinlan Shen, Yinfei Yang, Heming Ge, Daniel Cer, Gustavo~Hernández
  Ábrego, Keith Stevens, Noah Constant, Yun-Hsuan Sung, Brian Strope, and Ray
  Kurzweil. 2018.
\newblock \href {https://doi.org/10.18653/v1/w18-6317} {Effective {Parallel}
  {Corpus} {Mining} using {Bilingual} {Sentence} {Embeddings}}.
\newblock In \emph{Proceedings of the {Third} {Conference} on {Machine}
  {Translation}: {Research} {Papers}, {WMT} 2018, {Belgium}, {Brussels},
  {October} 31 - {November} 1, 2018}, pages 165--176. Association for
  Computational Linguistics.

\bibitem[{Hu et~al.(2020)Hu, Ruder, Siddhant, Neubig, Firat, and
  Johnson}]{hu_xtreme_2020}
Junjie Hu, Sebastian Ruder, Aditya Siddhant, Graham Neubig, Orhan Firat, and
  Melvin Johnson. 2020.
\newblock \href {http://arxiv.org/abs/2003.11080} {{XTREME}: {A} {Massively}
  {Multilingual} {Multi}-task {Benchmark} for {Evaluating} {Cross}-lingual
  {Generalization}}.
\newblock ArXiv:2003.11080 [cs].

\bibitem[{Huang et~al.(2019)Huang, Liang, Duan, Gong, Shou, Jiang, and
  Zhou}]{huang_unicoder_2019}
Haoyang Huang, Yaobo Liang, Nan Duan, Ming Gong, Linjun Shou, Daxin Jiang, and
  Ming Zhou. 2019.
\newblock \href {https://doi.org/10.18653/v1/D19-1252} {Unicoder: {A}
  {Universal} {Language} {Encoder} by {Pre}-training with {Multiple}
  {Cross}-lingual {Tasks}}.
\newblock In \emph{Proceedings of the 2019 {Conference} on {Empirical}
  {Methods} in {Natural} {Language} {Processing} and the 9th {International}
  {Joint} {Conference} on {Natural} {Language} {Processing}
  ({EMNLP}-{IJCNLP})}, pages 2485--2494, Hong Kong, China. Association for
  Computational Linguistics.

\bibitem[{Oord et~al.(2018)Oord, Li, and Vinyals}]{oord_representation_2018}
Aäron van~den Oord, Yazhe Li, and Oriol Vinyals. 2018.
\newblock \href {http://arxiv.org/abs/1807.03748} {Representation {Learning}
  with {Contrastive} {Predictive} {Coding}}.
\newblock \emph{CoRR}, abs/1807.03748.
\newblock ArXiv: 1807.03748.

\bibitem[{Schwenk(2018)}]{schwenk_filtering_2018}
Holger Schwenk. 2018.
\newblock \href {https://doi.org/10.18653/v1/P18-2037} {Filtering and {Mining}
  {Parallel} {Data} in a {Joint} {Multilingual} {Space}}.
\newblock In \emph{Proceedings of the 56th {Annual} {Meeting} of the
  {Association} for {Computational} {Linguistics}, {ACL} 2018, {Melbourne},
  {Australia}, {July} 15-20, 2018, {Volume} 2: {Short} {Papers}}, pages
  228--234. Association for Computational Linguistics.

\bibitem[{Tiedemann(2012)}]{opus}
Jorg Tiedemann. 2012.
\newblock Parallel {Data}, {Tools} and {Interfaces} in {OPUS}.
\newblock \emph{In Proceedings of the 8th International Conference on Language
  Resources and Evaluation (LREC'2012)}.

\bibitem[{Vaswani et~al.(2017)Vaswani, Shazeer, Parmar, Uszkoreit, Jones,
  Gomez, Kaiser, and Polosukhin}]{vaswani_attention_2017}
Ashish Vaswani, Noam Shazeer, Niki Parmar, Jakob Uszkoreit, Llion Jones,
  Aidan~N Gomez, Łukasz Kaiser, and Illia Polosukhin. 2017.
\newblock \href
  {https://proceedings.neurips.cc/paper/2017/hash/3f5ee243547dee91fbd053c1c4a845aa-Abstract.html}
  {Attention is {All} you {Need}}.
\newblock In \emph{Advances in {Neural} {Information} {Processing} {Systems}},
  volume~30. Curran Associates, Inc.

\bibitem[{Yang et~al.(2019)Yang, Ábrego, Yuan, Guo, Shen, Cer, Sung, Strope,
  and Kurzweil}]{yang_improving_2019}
Yinfei Yang, Gustavo~Hernández Ábrego, Steve Yuan, Mandy Guo, Qinlan Shen,
  Daniel Cer, Yun-Hsuan Sung, Brian Strope, and Ray Kurzweil. 2019.
\newblock \href {https://doi.org/10.24963/ijcai.2019/746} {Improving
  {Multilingual} {Sentence} {Embedding} using {Bi}-directional {Dual} {Encoder}
  with {Additive} {Margin} {Softmax}.}
\newblock In \emph{Proceedings of the {Twenty}-{Eighth} {International} {Joint}
  {Conference} on {Artificial} {Intelligence}, {IJCAI} 2019, {Macao}, {China},
  {August} 10-16, 2019}, pages 5370--5378.

\bibitem[{Yang et~al.(2021)Yang, Yang, Cer, Law, and
  Darve}]{yang_universal_2021}
Ziyi Yang, Yinfei Yang, Daniel Cer, Jax Law, and Eric Darve. 2021.
\newblock \href {https://doi.org/10.18653/v1/2021.emnlp-main.502} {Universal
  {Sentence} {Representation} {Learning} with {Conditional} {Masked} {Language}
  {Model}}.
\newblock In \emph{Proceedings of the 2021 {Conference} on {Empirical}
  {Methods} in {Natural} {Language} {Processing}, {EMNLP} 2021, {Virtual}
  {Event} / {Punta} {Cana}, {Dominican} {Republic}, 7-11 {November}, 2021},
  pages 6216--6228. Association for Computational Linguistics.

\bibitem[{Zweigenbaum et~al.(2017)Zweigenbaum, Sharoff, and Rapp}]{bucc}
Pierre Zweigenbaum, Serge Sharoff, and Reinhard Rapp. 2017.
\newblock \href {https://doi.org/10.18653/v1/w17-2512} {Overview of the
  {Second} {BUCC} {Shared} {Task}: {Spotting} {Parallel} {Sentences} in
  {Comparable} {Corpora}}.
\newblock In \emph{Proceedings of the 10th {Workshop} on {Building} and {Using}
  {Comparable} {Corpora}, {BUCC}@{ACL} 2017, {Vancouver}, {Canada}, {August} 3,
  2017}, pages 60--67. Association for Computational Linguistics.

\end{thebibliography}
\bibliographystyle{acl_natbib}

\appendix

\section{Full Tatoeba Results}\label{app:tatoeba}
We report the Tatoeba retrieval accuracy of all 36 languages in Table~\ref{tab:full_tatoeba_en} and Table~\ref{tab:full_tatoeba_xx}. Our approach consistently outperforms other baselines in both directions for most languages, with the advantage being particularly significant in the "en$\to$xx" direction. We observed that the performance of the TR-only model can vary much between the two directions, as demonstrated by languages such as jv, kk, sw, and tl. In contrast, our approach exhibits much more stable performance, which is beneficial for bidirectional applications.

\setlength\tabcolsep{2.5pt}
\begin{table*}[t]
    \small
    \centering
    \begin{tabular}{l|cccccccccccccccccc}
         \toprule
         Model & af & ar & bg & bn & de & el & es & et & eu & fa & fi & fr & he & hi & hu & id & it & ja\\
         \hline
         mBERT+TR & 95.5 & 90.3 & 94.5 & 88.8 & 99.1 & 96.4 & 98.1 & 97.4 & 95.1 & 94.0 & 96.5 & 95.3 & 90.6 & 95.5 & 96.8 & 95.4 & 94.4 & 96.1\\
         mBERT+TR+TLM & 95.9 & 89.7 & 94.6 & 87.0 & 99.1 & 95.6 & 98.3 & 96.8 & 95.0 & 94.0 & 95.7 & 95.4 & 91.5 & 95.6 & 95.6 & 95.0 & 93.8 & 95.0\\
         \textbf{mBERT+DAP} & 96.9 & 91.8 & 95.4 & 89.3 & 99.1 & 96.8 & 98.4 & 98.0 & 96.2 & 95.9 & 97.1 & 95.5 & 93.0 & 96.8 & 97.0 & 95.9 & 95.5 & 96.7\\
         \hline
         XLM-R+TR & 95.0 & 90.0 & 92.9 & 89.3 & 99.1 & 93.9 & 98.1 & 97.8 & 95.3 & 95.3 & 96.9 & 95.3 & 91.1 & 96.4 & 97.0 & 95.1 & 94.4 & 96.1\\
         XLM-R+TR+TLM & 92.7 & 90.2 & 94.3 & 88.8 & 99.1 & 95.5 & 97.3 & 96.8 & 93.8 & 94.4 & 95.9 & 94.2 & 91.2 & 96.4 & 95.9 & 96.0 & 94.4 & 94.2\\
         \textbf{XLM-R+DAP}& 96.1 & 93.1 & 95.7 & 91.4 & 99.2 & 96.7 & 98.4 & 98.1 & 96.0 & 94.9 & 97.3 & 95.5 & 93.6 & 97.3 & 97.0 & 96.4 & 96.3 & 96.2\\
         \hline
          & jv & ka & kk & ko & ml & mr & nl & pt & ru & sw & ta & te & th & tl & tr & ur & vi & zh\\
         \hline
         mBERT+TR & 29.3 & 81.0 & 62.6 & 91.2 & 97.7 & 91.6 & 96.2 & 95.4 & 95.6 & 75.1 & 84.0 & 90.2 & 96.2 & 67.7 & 98.2 & 89.6 & 96.9 & 95.3\\
         mBERT+TR+TLM & 31.2 & 79.2 & 64.7 & 91.8 & 97.5 & 92.0 & 95.9 & 95.4 & 94.8 & 77.2 & 85.3 & 89.7 & 96.0 & 71.0 & 97.7 & 91.3 & 96.9 & 95.3\\
         \textbf{mBERT+DAP} & 30.2 & 79.9 & 63.8 & 93.2 & 98.5 & 92.5 & 96.6 & 96.2 & 95.5 & 77.9 & 83.1 & 88.5 & 96.9 & 70.1 & 98.5 & 90.8 & 97.5 & 95.4\\
         \hline
         XLM-R+TR & 46.3 & 90.5 & 75.7 & 92.7 & 98.5 & 93.2 & 96.7 & 95.4 & 94.7 & 73.3 & 84.4 & 93.6 & 96.7 & 74.2 & 97.2 & 91.6 & 97.5 & 95.7\\
         XLM-R+TR+TLM & 23.4 & 92.4 & 69.2 & 91.6 & 97.2 & 90.4 & 95.7 & 95.5 & 94.3 & 72.8 & 71.0 & 88.5 & 96.4 & 55.8 & 97.1 & 85.9 & 97.0 & 94.6\\
         \textbf{XLM-R+DAP}& 27.3 & 93.7 & 68.5 & 93.3 & 98.4 & 92.5 & 96.6 & 96.1 & 95.4 & 77.2 & 80.8 & 92.3 & 98.2 & 65.6 & 98.3 & 90.3 & 98.2 & 95.4\\
         \bottomrule
    \end{tabular}
    \caption{Retrieval accuracy on 36 languages of direction xx$\to$en.}
    \label{tab:full_tatoeba_en}
\end{table*}
\begin{table*}[t]
    \small
    \centering
    \begin{tabular}{l|cccccccccccccccccc}
         \toprule
         Model & af & ar & bg & bn & de & el & es & et & eu & fa & fi & fr & he & hi & hu & id & it & ja\\
         \hline
         mBERT+TR & 94.8 & 88.7 & 93.3 & 86.2 & 98.8 & 95.4 & 97.4 & 96.3 & 94.7 & 94.3 & 95.6 & 95.8 & 89.7 & 95.0 & 95.6 & 94.3 & 95.1 & 95.9\\
         mBERT+TR+TLM & 95.7 & 88.0 & 93.8 & 85.8 & 98.9 & 96.1 & 97.6 & 96.3 & 94.8 & 93.7 & 94.8 & 95.3 & 89.6 & 95.3 & 94.4 & 94.1 & 94.1 & 95.3\\
         \textbf{mBERT+DAP} & 96.3 & 90.6 & 94.3 & 87.8 & 98.9 & 96.1 & 98.1 & 98.0 & 96.0 & 95.6 & 96.4 & 95.4 & 92.2 & 96.0 & 96.5 & 95.2 & 95.8 & 96.6\\
         \hline
         XLM-R+TR & 87.6 & 90.3 & 92.0 & 85.5 & 98.3 & 95.9 & 96.2 & 95.9 & 92.8 & 93.1 & 95.4 & 92.4 & 91.6 & 94.3 & 95.6 & 94.0 & 94.4 & 90.9\\
         XLM-R+TR+TLM & 96.1 & 89.3 & 93.9 & 90.0 & 99.1 & 93.9 & 98.2 & 97.0 & 94.9 & 95.7 & 96.8 & 95.4 & 89.6 & 97.1 & 96.5 & 95.3 & 94.4 & 96.4\\
         \textbf{XLM-R+DAP}& 96.3 & 92.2 & 95.4 & 91.2 & 98.9 & 96.6 & 98.6 & 98.1 & 95.7 & 96.0 & 97.1 & 96.3 & 93.1 & 97.0 & 97.2 & 96.3 & 96.1 & 97.3\\
         \hline
          & jv & ka & kk & ko & ml & mr & nl & pt & ru & sw & ta & te & th & tl & tr & ur & vi & zh\\
         \hline
         mBERT+TR & 43.4 & 81.5 & 66.4 & 91.8 & 97.4 & 92.3 & 96.1 & 94.6 & 94.8 & 72.3 & 83.4 & 89.3 & 95.8 & 70.6 & 96.8 & 89.5 & 97.3 & 94.3\\
         mBERT+TR+TLM & 46.3 & 78.0 & 67.8 & 92.5 & 98.0 & 92.2 & 95.9 & 94.7 & 94.2 & 74.9 & 84.0 & 89.7 & 95.8 & 74.6 & 96.8 & 90.4 & 97.6 & 94.9\\
         \textbf{mBERT+DAP} & 47.3 & 80.8 & 65.4 & 92.3 & 98.3 & 93.3 & 97.2 & 95.6 & 94.8 & 75.6 & 82.4 & 89.7 & 96.4 & 75.5 & 98.2 & 91.7 & 97.8 & 95.3\\
         \hline
         XLM-R+TR & 16.1 & 88.3 & 57.6 & 89.8 & 96.2 & 87.3 & 95.4 & 95.5 & 93.9 & 59.5 & 62.5 & 81.6 & 95.3 & 46.8 & 97.0 & 82.2 & 96.7 & 92.8\\
         XLM-R+TR+TLM & 49.8 & 90.6 & 82.6 & 92.4 & 98.5 & 94.2 & 97.0 & 95.0 & 94.2 & 81.5 & 86.0 & 96.6 & 96.9 & 80.2 & 96.6 & 92.6 & 97.7 & 95.2\\
         \textbf{XLM-R+DAP}& 47.3 & 91.6 & 75.3 & 93.4 & 99.0 & 93.6 & 96.8 & 95.6 & 95.1 & 78.5 & 86.3 & 94.9 & 97.8 & 77.1 & 97.9 & 92.7 & 98.0 & 96.0\\
         \bottomrule
    \end{tabular}
    \caption{Retrieval accuracy on 36 languages of direction en$\to$xx.}
    \label{tab:full_tatoeba_xx}
\end{table*}

\section{Scoring Function For BUCC}\label{app:bucc}
In contrast to direction comparison between similarities, margin-based method accounts for the scale inconsistencies of measure. We adopted the method proposed by~\citet{artetxe_margin-based_2019}:
\begin{equation}
    f(x,y)=\frac{\phi(x,y)}{\sum_{z\in N_k(x)}\frac{\phi(x,y)}{k}+\sum_{z\in N_k(y)}\frac{\phi(z, y)}{k}},
\end{equation}
where $N_k(x)$ denotes the set of $k$ nearest neighbours of $x$ in the other language. In our experiments, we set $k=4$.

With a certain threshold $\gamma$, sentence pairs such that $f(x,y)\geq\gamma$ are identified as aligned. For those $x$ appearing in multiple aligned pairs, we select the pair with the highest score.

To decide the best threshold, we first compute the scores of all candidates and sort them into an ordered sequence. Next, we compute F1 score by setting $\gamma$ to each middle point of two consecutive scores and find the optimal $\gamma$. This procedure is done on training set.

\section{XNLI Fine-tuning}\label{app:xnli}
The fine-tuning hyperparamter setting is shown in Table~\ref{tab:hyper_xnli}. We searched the learning rate among \{1e-5, 3e-5, 5e-5, 7e-5\}.

\begin{table}[h]
    \centering
    \begin{tabular}{lc}
    \toprule
        Batch size & 256 \\
        Learning rate & 5e-5 \\
        Epochs & 2 \\
        Max seq length & 128 \\
        Weight decay & 0 \\
    \bottomrule
    \end{tabular}
    \caption{Hyperparameter setting of XNLI fine-tuning.}
    \label{tab:hyper_xnli}
\end{table}

\end{document}